\documentclass[10pt,twocolumn,letterpaper]{article}

\usepackage{iccv}
\usepackage{times}
\usepackage{epsfig}
\usepackage{graphicx}
\usepackage{amsmath}
\usepackage{subcaption}
\usepackage{amssymb}
\usepackage{booktabs}
\usepackage{dblfloatfix}
\usepackage[sort]{cite}
\usepackage[font=footnotesize,subrefformat=parens]{subcaption}
\usepackage{tikz}
\usetikzlibrary{fit}

\newcommand{\refsec}[1]{Section~\ref{sec:#1}}

\newcommand{\refeq}[1]{Eq.~\ref{eq:#1}}

\newcommand{\lblfig}[1]{\label{fig:#1}}
\newcommand{\lblsec}[1]{\label{sec:#1}}
\newcommand{\lbleq}[1]{\label{eq:#1}}

\renewcommand{\vec}{\boldsymbol}

\usepackage[pagebackref=true,breaklinks=true,letterpaper=true,colorlinks,bookmarks=false]{hyperref}

 \iccvfinalcopy %

\def\iccvPaperID{372} %

\ificcvfinal\fi
\begin{document}

\title{Learning Data-driven Reflectance Priors \\
for Intrinsic Image Decomposition}

\author{Tinghui Zhou\\
UC Berkeley\\
\and
Philipp Kr\"ahenb\"uhl\\
UC Berkeley\\
\and
Alexei A. Efros\\
UC Berkeley
}

\maketitle

\begin{abstract}

We propose a data-driven approach for intrinsic image decomposition, which is the process of inferring the confounding factors of reflectance and shading in an image. We pose this as a two-stage learning problem. First, we train a model to predict relative reflectance ordering between image patches (`brighter', `darker', `same') from large-scale human annotations, producing a data-driven reflectance prior. Second, we show how to naturally integrate this learned prior into existing energy minimization frameworks for intrinsic image decomposition. We compare our method to the state-of-the-art approach of Bell~\etal~\cite{bbs-iiw-14} on both decomposition and image relighting tasks, demonstrating the benefits of the simple relative reflectance prior, especially for scenes under challenging lighting conditions.
\end{abstract}

\section{Introduction}
\begin{figure}[t]
\centering
\includegraphics[width=1\linewidth]{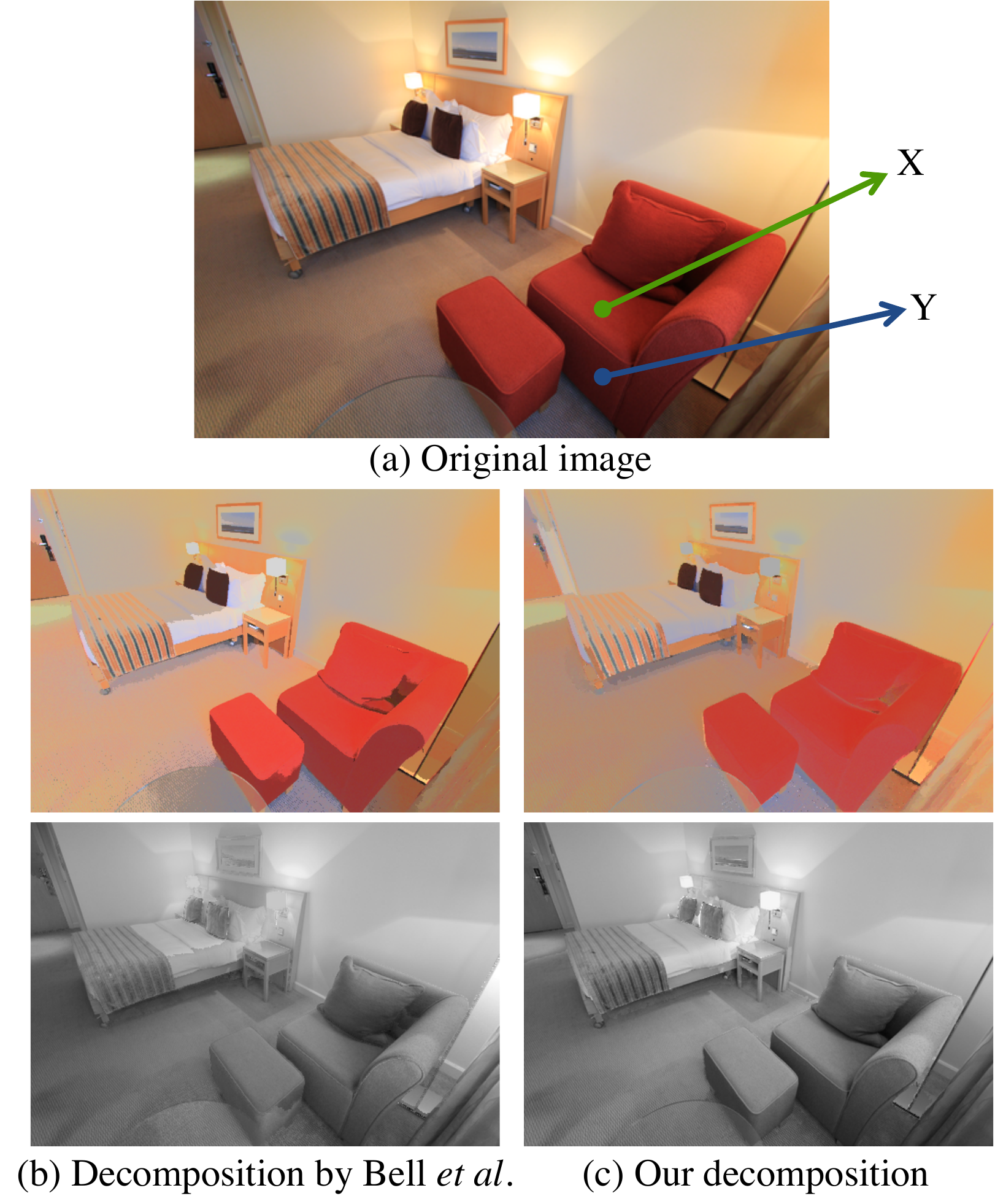}
\caption{Given an image (a), people have no trouble
disentangling the confounding factors of reflectance and shading:  we can see that $X$ is much brighter than $Y$, but {\em at the same time}, we can also clearly tell that they are both ``made of the same stuff'' and have the same surface reflectance. Our algorithm (c) automatically decomposes (a) into a reflectance image (c,top) and a shading image (c,bottom). Note how the mini-sofa is a uniform red in our reflectance image, compared to (b) state-of-the-art algorithm of Bell~\etal~\cite{bbs-iiw-14}.}
\label{fig:teaser}
\end{figure}

The human visual system is remarkable in its ability to decompose the jumbled mess of confounds that is our visual world into simpler underlying factors.  Nowhere is this more apparent than in our impressive ability, even from a single still image, to tease apart the effects of surface reflectance vs. scene illumination.  Consider the mini-sofa in Figure~\ref{fig:teaser}(a): on one hand, we can see that its seat (point $X$) is much brighter than its frontal face (point $Y$), but {\em at the same time}, we can also clearly tell that they are both ``made of the same stuff'' and have the same surface reflectance. This is remarkable because, by the time the light has bounced off the sofa toward the eye (or the camera), the contributions of reflectance and illumination have been hopelessly entangled, which the brain then needs to undo. 

In computer vision, the decomposition of an image into reflectance (albedo) and illumination (shading) maps is usually, if somewhat inaccurately, referred to as the {\em intrinsic image decomposition}~\cite{bt78}\footnote{The original formulation of Barrow and Tenenbaum~\cite{bt78} also includes other factors, such as depth, orientation, occlusion, transparency, etc}.  The intrinsic image model states that the observed luminance image is the product of the reflectance image times the shading image.  Clearly, the inverse problem of inferring the underlying reflectance and shading images is ill-posed and under-constrained in this pure form since any given pixel intensity could be explained equally well by reflectance or shading~\cite{adelson96}.  To address this, additional constraints (priors) are typically imposed on the decomposition process to capture the statistical and/or physical regularities in natural images. However, those priors are typically hand-crafted and overly weak.
For example, one popular prior proposed originally in the  Retinex algorithm of Land and McCann~\cite{retinex} assumes that large intensity gradients correspond to reflectance edges, while low-frequency changes are mainly due to shading.  While this prior works well in many cases, it fails in the presence of strong shadows, sharp changes in surface orientation, and smoothly-varying planar textures. 
Since then, many other clever priors have been proposed, including texture statistics~\cite{oh01,liu12}, shape, albedo, and illumination~\cite{barron12a,barron12b,barron13}, meso- and macro-scales of shading~\cite{liao13}, chromaticity segmentation~\cite{garces12}, sparsity on reflectances~\cite{omer04,gehler11}, etc., or combination thereof~\cite{bbs-iiw-14},
in the hopes of finding the silver bullet which could {\em fully explain} the intrinsic image phenomenon, but to date none has emerged.  One is faced with the possibility that there might not exist a simple, analytic prior and that a more data-driven approach is warranted.

In this paper we propose to learn priors for intrinsic image decomposition directly from data.  Compared to other work that trains a reflectance vs. shading classifier on image patches (e.g.~\cite{tappen05,tappen06}), our main contribution is to train a {\em relative} reflectance prior on {\em pairs} of patches.  Intuitively, the goal is to learn to detect surface regions with similar reflectance, even when their intensities are different.  We take advantage of the recently released Intrinsic Images in the Wild (IIW) database of Bell~\etal~\cite{bbs-iiw-14}, in which a large set of relative reflectance judgments are collected from human subjects for a variety of real-world scenes. Other contemporary work, developed independently, have also employed the IIW dataset. Narihira~\etal~\cite{takuya-cvpr15} use the IIW dataset to learn a perceptual lightness model. The key difference is that we not only learn a relative reflectance prior from pairwise annotations, but also utilize it for intrinsic image decomposition. In these same proceedings, Zoran~\etal~\cite{zoran15} use a similar approach to ours to estimate ordinal relationships between pairs of points, but globalizes them with a different energy optimization.

Our relative reflectance model is an end-to-end trained convolutional neural network that predicts a probability distribution over the relative reflectance (`brighter', `darker', `same') between two query pixels. We show how to naturally integrate this learned prior into existing energy minimization frameworks for intrinsic image decomposition, and demonstrate the benefits of such relative reflectance priors, especially for scenes under challenging illumination conditions.

\section{Learning a model of reflectance}

Let $r_i \in \mathcal{R}$ be a reflectance estimate at pixel $i$, where $\mathcal{R}$ is the set of all reflectance values in a scene.
For two reflectance values $r_i,r_j \in \mathcal{R}$ let $r_i<r_j$ denote that reflectance $r_i$ is darker than reflectance $r_j$, and $r_i=r_j$ means that the reflectances are roughly equivalent.

Estimating reflectance directly is hard and usually requires a specialized sensor, such as a photometer.
Not even the human visual system can infer absolute reflectance reliably (see Adelson~\cite{ted} for examples).
Humans are much better at estimating relative reflectance between two point $r_i$ and $r_j$ in a scene~\cite{bbs-iiw-14}.
We follow this intuition and learn a classifier that predicts this relative reflectance between different parts of a scene in \refsec{rel_light}.
However, just like human reflectance estimates, this classifier might not be globally consistent.
\refsec{abs_light} recovers the globally consistent reflectance estimate following our relative estimates.
We then use this global reflectance model in \refsec{decomp} to guide an intrinsic image decomposition.

\subsection{Relative reflectance classifier}
\lblsec{rel_light}
For two pixels $i$ and $j$ in a scene, our goal is to estimate the relative reflectance between them as being equal $r_{i} = r_{j}$, darker $r_{i} < r_{j}$ or brighter $r_{i} > r_{j}$. Our relative reflectance classifier is a multi-stream convolutional neural network (see Fig.~\ref{fig:net}), accounting for 1) local features around pixel $i$, 2) local features around pixel $j$, 3) global scene features of the input image, and 4) spatial coordinates of both input pixels, respectively. The network weights are shared between the two local feature extraction streams. All features are then concatenated, and fed through three fully-connected layers that predict classification scores over the relative reflectance labels (`same', `darker', `brighter'). Each convolution and fully-connected layer (except for the last prediction layer) is followed by a rectified linear unit.

\begin{figure}[t]
\centering
\includegraphics[width=\linewidth]{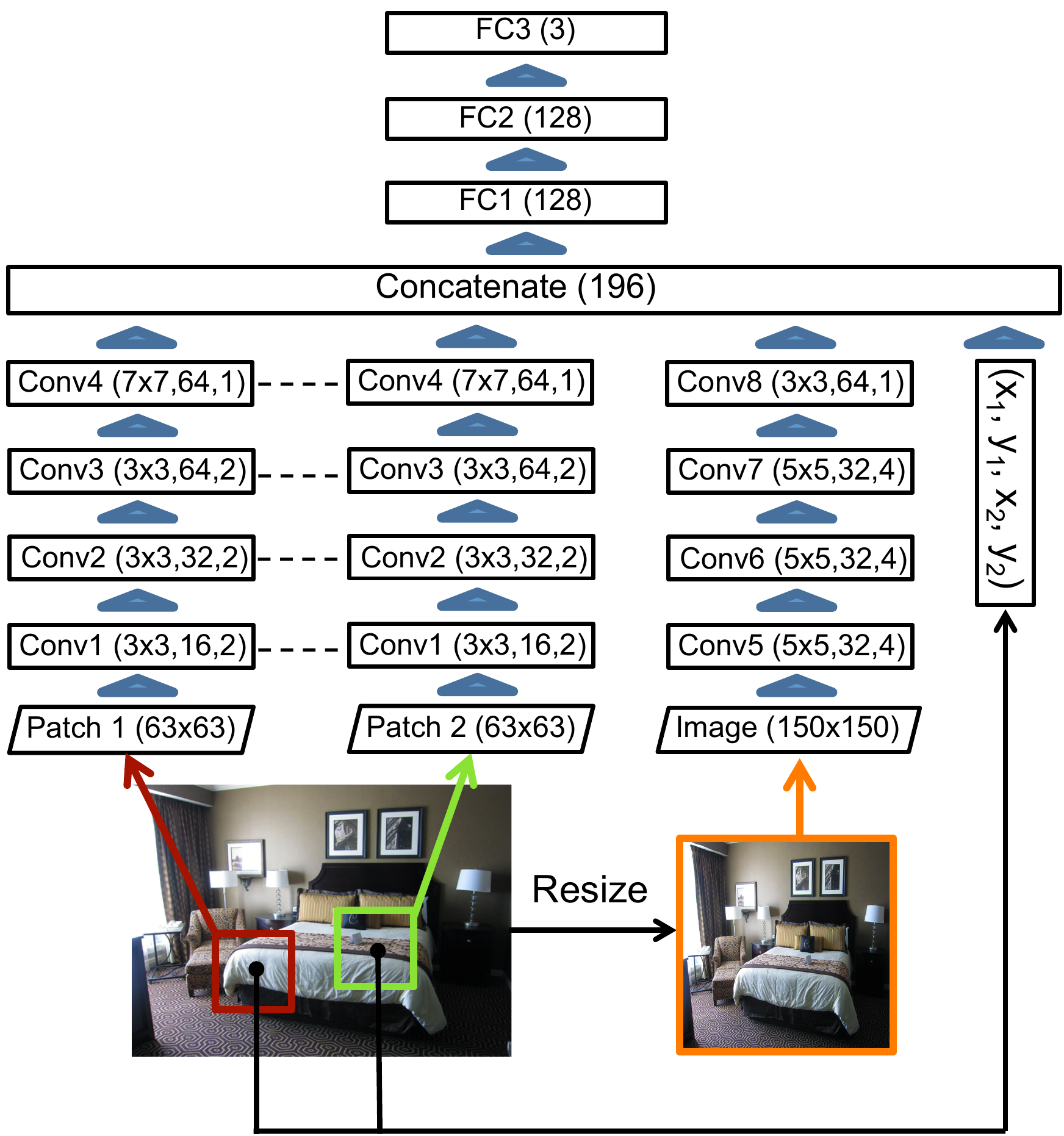}
\caption{Our multi-stream network architecture for relative reflectance prediction. The network weights are shared between the local feature extraction streams. Features extracted from all four streams are fed through three fully-connected layers for final relative reflectance prediction (see Sec.~\refsec{rel_light} for more details). }
\lblfig{net}
\end{figure}

We train this network from scratch using the pairwise human judgments of the Intrinsic Images in the Wild dataset~\cite{bbs-iiw-14} and millions more obtained through symmetry and transitivity properties of the original annotations (see \refsec{dataaug} for details on data augmentation). The network is learned end-to-end in {\sc Caffe} \cite{jsdkl-caffe-14} using a softmax loss. Our network outperforms all state-of-the-art methods in terms of relative reflectance predictions, as we will show in \refsec{results}. However the resulting predictions are not always globally consistent. This is in part due to inconsistencies in the human-annotated training data. Roughly $7.5\%$ of all training annotations are inconsistently labeled~\cite{bbs-iiw-14} and our network learns part of that inconsistency.

Next, we show how to recover a globally consistent reflectance estimate from the noisy pairwise predictions produced by the classifier.

\subsection{Globally consistent reflectance estimate}
\lblsec{abs_light}
The network output gives an estimate for the relative reflectance between a pair of pixels $i$ and $j$.
Let $w_{=,i,j}$, $w_{<,i,j}$ and $w_{>,i,j}$ be the classifier score of `same', `darker', and `brighter', respectively.
A higher score enforces a larger consistency for a specific pairwise comparison.
We constrain all weights to be non-negative, and formulate global reflectance estimation as a constrained optimization problem, where each classifier output imposes a pairwise constraint on the global ordering
\begin{align}
\underset{\vec r,\varepsilon}{\text{minimize}} & & & \sum_{i,j \in \mathcal{E}} \ \sum_{o\in\{=,<,>\}} w_{o,j,i} \xi_{o,i,j} & \notag\\
\text{subject to} & & & r_i\!\le\!r_j\!+\!\xi_{=,i,j} &\notag\\
  & & & r_j \le r_i + \xi_{=,i,j}, &\notag\\
  & & & r_i \le r_j + \xi_{<,i,j}, &\notag\\
  & & & r_j \le r_i + \xi_{>,i,j}, &\notag\\
  & & & \xi \ge 0.\lbleq{emin}
\end{align}
Here, $\xi$ is a slack variable that tries to enforce all constraints as well as possible.
All pairwise reflectance measures are evaluated on a set of sparse edges $\mathcal{E}$.

This constrained optimization naturally translates into a global energy minimization:
\begin{equation}
 E(\vec x) = \sum_{i,j \in \mathcal{E}} \ \sum_{o\in\{=,<,>\}} w_{o,j,i} \ \mu_{o}(r_i,r_j) \lbleq{energy_sparse},
\end{equation}
where $\mu_{<}$, $\mu_{>}$, and $\mu_{=}$ penalizes the disagreement between our classifier and the globally consistent ranking.

For objective \refeq{emin} this translates into a hinge loss, that penalizes the degree to which the consistent reflectance estimate disagrees with our classifier:
\begin{align*}
 \mu_{=}(r_i,r_j) &= \xi_{=,i,j} = |r_i - r_j|\\
 \mu_{<}(r_i,r_j) &= \xi_{<,i,j} = \max(r_i - r_j, 0)\\
 \mu_{>}(r_i,r_j) &= \xi_{>,i,j} = \max(r_j - r_i, 0).
\end{align*}
For continuous values $r_i$ the energy minimization \ref{eq:energy_sparse} is convex.
For discrete values $r_i$ it can be expressed as a binary submodular problem on an extended sparsely connected graph~\cite{kvt-imtcm-11}.
We use GraphCuts to globally optimize it~\cite{bk-ecmae-04}.

While objective \ref{eq:emin} computes a global ordering on the reflectance, it does not provide information about the absolute reflectance in an image.
In the next section we will show how to incorporate the reflectance prior into a standard intrinsic image decomposition pipeline to recover an absolute estimate of reflectance.

\section{Intrinsic image decomposition}
\lblsec{decomp}
We start out with the intrinsic image decomposition framework of Bell~\etal~\cite{bbs-iiw-14}.
Given an input image $I$, their system recovers a reflectance image $\vec r$ and shading image $\vec s$.
They model intrinsic image decomposition as an energy minimization in a fully connected CRF~\cite{kk-eifcc-11}.
\begin{equation}
 E(\vec s,\vec r) =\!\sum_{i}\!\psi_i(r_i,s_i) +\!\sum_{i>j}\!\psi^{r}_{ij}(r_i,r_j) + \psi^{s}_{ij}(r_i,r_j),
\end{equation}
where $\psi_i$ is a unary term that captures some lightweight unary priors on absolute shading intensity or chromaticity of the reflectance as an L1 norm between the original image and the estimated properties.
The unary term also constrains the reflectance and shading to reconstruct the original image.
Most of the heavy lifting of the model is done by the pairwise terms $\psi^{r}$ and $\psi^{s}$ that enforce smoothness of reflectance and lighting respectively.

The pairwise shading term is modeled as a fully connected smoothness prior:
$$
\psi^{s}_{ij}(r_i,r_j) = (s_i - s_j)^2 \exp\left(-\beta_1(\vec p_i\!-\!\vec p_j)^2\right),
$$
where $\vec p_i$ is the position of a pixel $i$, and $\beta_1$ is a parameter controlling the spatial extent of the prior.
This prior captures the intuition that the shading varies smoothly over smooth surfaces.

The pairwise reflectance term is modeled as a color sensitive regularizer encouraging pixels with a similar color value in the original image to take a similar reflectance:
$$
\psi^{r}_{ij}(r_i,r_j) = |r_i - r_j| \exp\left(-\beta_2(\vec p_i\!-\!\vec p_j)^2 -\beta_3(\vec I_i\!-\!\vec I_j)^2\right),
$$
where $\vec I_i$ is color value of a pixel $i$, and $\beta_2$ and $\beta_3$ control the spatial and color extent of the prior.
This reflectance term is quite arbitrary, as original color values are usually not a good sole predictor of reflectance.
In the rest of this section we will show how to replace this term with our data-driven pairwise reflectance prior.

The overall energy $E(\vec s,\vec r)$ is optimized using an alternating optimization for $\vec s$ and $\vec r$.
The reflectance term $\vec r$ is optimized using the mean-field inference algorithm of Kr\"ahenb\"uhl and Koltun~\cite{kk-eifcc-11}, while the shading term is optimized with iteratively reweighted least squares (IRLS).

\subsection{Data-driven reflectance prior}
We now show how to incorporate our relative reflectance classifier into the mean-field inference for reflectance.
Specifically we define our new pairwise term as
\begin{equation}
 \psi^{r}_{ij}(r_i,r_j) = \sum_{o \in \{=,<,>\}} \mu_{o,i,j}(r_i,r_j) w_{o,i,j},
\end{equation}
The main difficulty here is to evaluate the pairwise term densely over the image.
The mean-field inference algorithm relies on an efficient evaluation of $\tilde Q_i(r_i) = \sum_{j}\sum_{r_j} \psi^{r}_{ij}(r_i,r_j)Q_(r_i)$, which is known as message passing.
This message passing step naturally decomposes into a matrix multiplication with $\mu_{o}$ and a filtering term with $w_{o}$.
The matrix multiplication can be evaluated efficiently as it is independent for each pixel and scales linearly in the number of pixels.
The filtering step on the other hand requires an exchange of information between each pair of pixels in the image.
Kr\"ahenb\"uhl and Koltun~\cite{kk-eifcc-11} showed that for a Gaussian pairwise term the filter can be approximated efficiently.
The same Gaussian pairwise term is used in the original model of Bell~\etal~\cite{bbs-iiw-14}.
In our model this filter is no longer a simple Gaussian kernel, but guided by the output of a classifier.
The filtering has the following form
\begin{equation}
 \hat Q^{(o)}_i(l) = \sum_j w_{o,i,j} Q_j(l), \lbleq{filter}
\end{equation}
for each comparison $o \in \{<,>,=\}$.
For our data-driven pairwise term we would need to evaluate a classifier densely over each pair of pixels in the image, which is computationally intractable for more than a few thousand pixels.

However, the classifier output is quite low rank. If we denote $|\mathcal{R}|$ as the number of unique reflectance values in a scene, which is usually small~\cite{omer04,gehler11}, then the output of an ideal classifier is of at most rank $|\mathcal{R}|$. This comes from the fact that each reflectance value $r \in \mathcal{R}$ forms a binary basis $B$, with a value of $B_{i,r}=1$ if pixel $i$ takes reflectance $r$, and $B_{i,r}=0$ otherwise. Thus any ideal classifier output can be expressed as a product of $B \tilde W_o B^\top$, where $\tilde W_o$ is a $|\mathcal{R}| \times |\mathcal{R}|$ matrix describing the weighting between different reflectance values.
Any rank beyond this can be attributed to noise or inconsistencies in the classifier.
We measured the rank of the classifier matrix by randomly sampling $K=500$ points in the image and computing the full pairwise term between those points.
This results in a $K \times K$ pairwise comparison matrix. We never encountered this classifier matrix to be of rank more than $100$.
This suggests that the low rank approximation models $w_{<}$, $w_{=}$ and $w_{>}$ well.

\subsection{Nystr{\"o}m approximation}
We use Nystr{\"o}m's method~\cite{nystrom} to approximate $w_{o}$.
The main caveat with Nystr{\"o}m is that it requires a symmetric pairwise comparison matrix $w_o$.
While the equality constraint matrix $w_{=}$ is symmetric, the inequality matrices are not $w_{>} = w_{<}^\top$.
We address this by rearranging all classifier outputs in a larger comparison matrix $W$:
$$
 W = \left( \begin{array}{cccccc}
w_{=,1,1} & w_{>,1,1} & w_{=,1,2} & w_{>,1,2} & \ldots \\
w_{<,1,1} & w_{=,1,1} & w_{<,1,2} & w_{=,1,2} & \ldots \\
w_{=,2,1} & w_{>,2,1} & w_{=,2,2} & w_{>,2,2} & \ldots \\
w_{<,2,1} & w_{=,2,1} & w_{<,2,2} & w_{=,2,2} & \ldots \\
\hdots & \hdots &\hdots &\hdots &
\end{array} \right)
$$
This extended matrix is symmetric and can be well approximated using Nystr{\"o}m's method.
It is still low rank, as the three submatrices it comprises of are all low rank.
We can compute the filtering in \refeq{filter} by multiplying $W$ with a vector $[Q_1(l),0,Q_2(l),0,Q_3(l),\ldots]^\top$ and extracting every other elements from it.

The Nystr{\"o}m approximation samples $2K$ rows from matrix $W$.
Let $C$ denote those sampled rows.
We always sample pairs of consecutive rows, to not introduce a bias towards any of the operations $=$,$<$ or $>$,
Nystr{\"o}m then approximates the dense pairwise classifier matrix as
$$
W \approx C D^+ C^\top,
$$
where $D$ is a $K \times K$ matrix corresponding to the dense pairwise classifier scores between all sampled points, and $^+$ refers to the pseudo-inverse.
We sample $K=64$ on a regular grid, which allows us to compute the matrices $C$ and $D$ within $10$ seconds including the classifier evaluation.
The Nystr{\"o}m approximation allows us to compute a message passing step within a few hundred milliseconds, while a naive evaluation would take multiple days to compute.

In summary, we evaluate the pairwise reflectance classifier from $K$ sampled points to all other points in the image.
The Nystr{\"o}m approximation then allows us to approximate a fully-connected dense pairwise comparison matrix using those few samples, which in turn allows for a natural integration into the fully connected CRF framework of Kr\"ahenb\"uhl and Koltun. Notice that Nystr{\"o}m approximation for dense CRF has recently been explored in~\cite{wang15}. However, \cite{wang15} merely approximates the commonly used Gaussian kernel, while we show how to integrate a more general output of a classifier into the dense CRF framework. 

\section{Experiments}
\lblsec{results}

In this section, we evaluate the performance of each component of our pipeline using two data sources: 1) Intrinsic Images in the Wild (IIW) dataset~\cite{bbs-iiw-14} and 2) Image Lighting Composition (ILC) dataset~\cite{webcam}. Our main baseline is the state-of-the-art intrinsic image decomposition algorithm by Bell~\etal~\cite{bbs-iiw-14}.
All models are trained and evaluated on the dataset split of Narihira~\etal~\cite{takuya-cvpr15}. 

\subsection{Data augmentation}
\lblsec{dataaug}
IIW dataset provides $875,833$ comparisons across $5,230$ photos, which we extensively augment by exploiting the symmetry and transitivity of the comparisons. The augmentation not only helps reduce overfitting (as shown in Sec.~\refsec{sec:np}), but also generates pixel pairs that are spatially distant from each other (in contrast to ones originally derived from edges of a Delauney triangulation~\cite{bbs-iiw-14}).  We create the augmented training and test annotations as follows:
\begin{enumerate}
\item Remove low-quality comparisons with human confidence score $< 0.5$.
\item For each remaining pairwise comparison $(r_i, r_j)$, augment the annotation for $(r_j, r_i)$ by either flipping (if $r_i \neq r_j$) or keeping (if $r_i = r_j$) the sign. 
\item For any unannotated pair of reflectances $(r_i, r_j)$ that share a comparison with $r_k$, we augment it using the following rules: 1) $r_i=r_j$, iff $r_i=r_k$ and $r_j=r_k$ for all connected $r_k$; 2) $r_i > r_j$, iff $r_i \ge r_k > r_j$ or $r_i > r_k \ge r_j$; 3) $r_i < r_j$, iff $r_i < r_k \le r_j$ or $r_i \le r_k < r_j$. If any pairwise comparisons are inconsistent we do not complete them. This step is done repetitively for each image until no further augmentation is possible. 
\end{enumerate}
Our augmentation generates $22,903,366$ comparisons in total, out of which $18,621,626$ are used for training and $4,281,740$ for testing.

\subsection{Network performance}
\lblsec{sec:np}
\begin{table}[t]                                             
\centering       
\begin{tabular}{ccc}  
\toprule
 Data source& Original & Augmented \tabularnewline
 Metric & WHDR & Error Rate \tabularnewline
 \midrule
 Bell \etal~\cite{bbs-iiw-14} & 20.6 & 27.9 \tabularnewline
 Retinex-Color~\cite{grosse09} & 26.9 & 29.3  \tabularnewline
 Retinex-Gray~\cite{grosse09} & 26.8  & 30.5 \tabularnewline
 Garces \etal~\cite{garces12} & 24.8  & 29.9 \tabularnewline
Shen and Yeo~\cite{shen11} & 32.5  & 34.2 \tabularnewline
Zhao \etal~\cite{zhao12}       & 23.8  & 31.1  \tabularnewline 
Narihira \etal~\cite{takuya-cvpr15}   & 18.1 & 36.6  \tabularnewline 
 \midrule
Local                           & 16.6  & 25.8  \tabularnewline
Local + Spatial                  & 16.1  & 25.1  \tabularnewline
Local + Spatial + Global         & $\mathbf{15.7}$ & $\mathbf{24.6}$   \tabularnewline
Local + Spatial + Global (Orig.) & 17.3 & 32.4 \tabularnewline
\bottomrule
\end{tabular}                                                  
\caption{Performance on the IIW dataset~\cite{bbs-iiw-14} measured by WHDR (left) on the original, locally-connected comparisons and Error rate (right) on our augmented, potentially long-range comparisons. The bottom four rows correspond to our models trained with different components: local features only, local and spatial features, full network, and full network trained on original IIW annotations only.}
\label{tab:err}
\vspace{-3mm}
\end{table}

We use {\sc Adam}~\cite{adam} with $\beta_1 = 0.9$, $\beta_2 = 0.999$, an initial learning rate of $0.001$, step size of $20,000$, a step multiplier $\gamma=0.8$. We train with mini-batches of $128$ pairs and weight decay of $0.002$. 

For evaluation, we first use the same \emph{weighted human disagreement rate} (WHDR) metric as~\cite{bbs-iiw-14} on the test split. WHDR measures the percent of human judgments that a model incorrectly predicts, weighted by the confidence of each judgment. Note that the human judgments are not necessarily consistent in the IIW dataset as human performance using this metric is 7.5~\cite{bbs-iiw-14}. As shown in Table~\ref{tab:err}, our full model trained on the augmented data performs the best with WHDR $= 15.7$.

Additionally, we evaluate the error rate of different algorithms on our augmented annotations. Our full model again obtains the lowest error rate of $24.6$. More surprisingly, on this metric other baselines surpass the recent top performer~\cite{takuya-cvpr15}. 
This is likely due to a subtle bias 
in the original IIW annotations -- 
spatially close pixels often have the same reflectance.
This bias is no longer present in our augmented annotations as they contain more long-range pairs. This is further verified by the performance of our full model trained only on the original annotations: it too does poorly on the augmented data.

\begin{figure*}[t]
\centering
\includegraphics[width=0.85\linewidth]{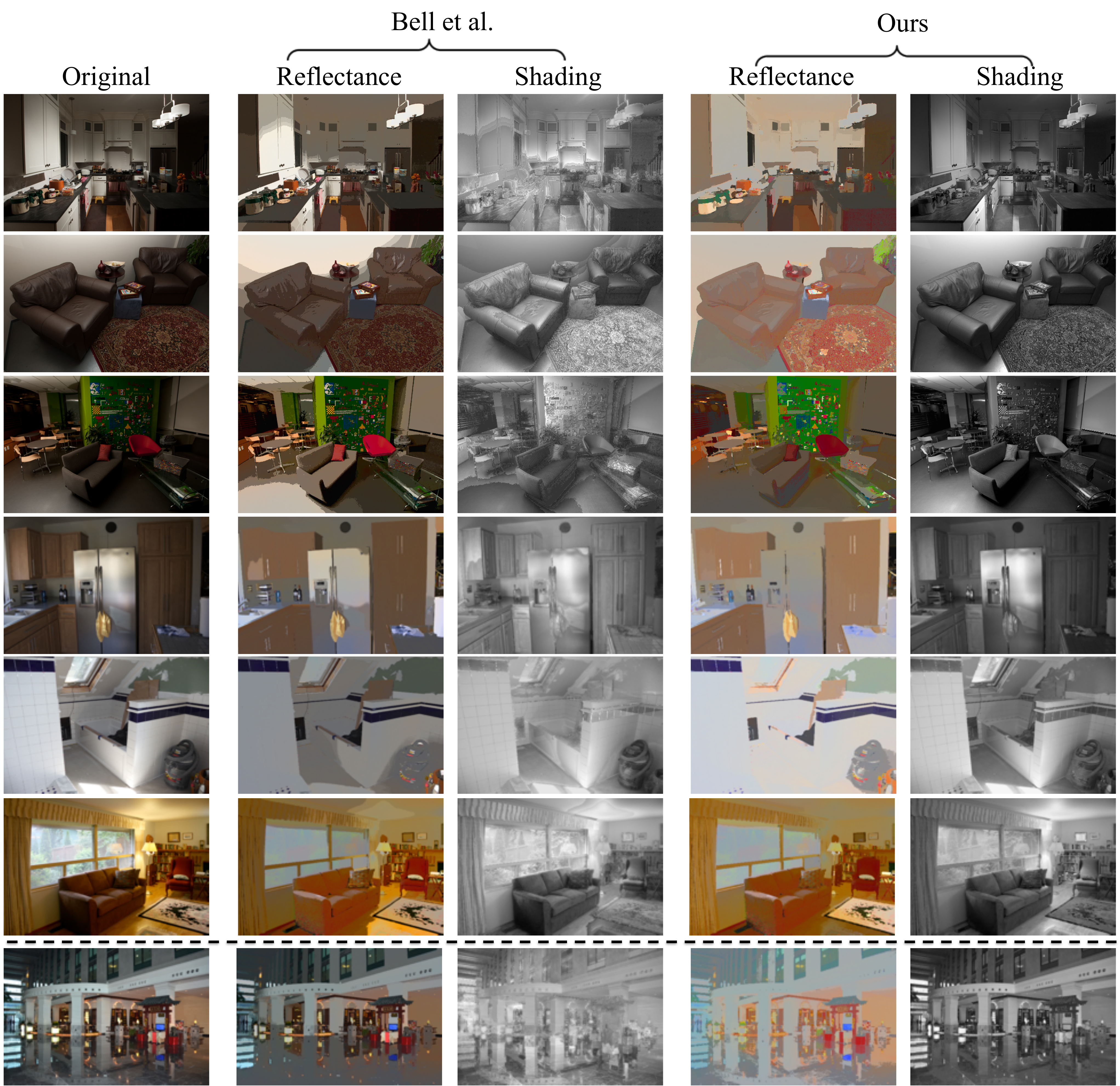}
\caption{Comparison of intrinsic image decomposition between Bell~\etal~\cite{bbs-iiw-14} and ours (chrom. + our prior + shading). Rows 1--3 are examples from the ILC dataset~\cite{webcam}, and the rest are ones from the IIW dataset~\cite{bbs-iiw-14}. In general, our decomposition tends to distinguish between reflectance and shading boundaries better compared to the baseline, especially under challenging lighting conditions (e.g. Rows 1--3). The last row shows an example where Bell \etal outperforms ours due to stronger reflectance smoothness constraints.}
\label{fig:decomp}
\vspace{-3mm}
\end{figure*}

\paragraph{Globally consistent reflectance estimate}
We measure the performance of recovering a globally consistent reflectance estimate with the energy optimization presented in Sec.~\ref{sec:abs_light}. Specifically, for each test image in the IIW dataset, we build a sparse graph over the annotated pixel pairs, and apply the relative reflectance network to each of the sampled pixels. The predicted scores are then jointly optimized by Eq.~\ref{eq:energy_sparse} using GraphCuts~\cite{bk-ecmae-04} to recover the globally consistent ordering. The recovery performance is measured using WHDR, and we obtain $18.0$ over the entire test split. Compared to the direct network output (WHDR $= 15.7$), global ordering recovery loses $2.3$ percent of the performance due to the inconsistency and noise of the network output. 

\paragraph{Nystr{\"o}m approximation}
We experimented with different point sampling strategies (including random sampling, spatial grid sampling and Poisson disk sampling) as well as different sample sizes, and found that grid sampling with $64$ samples to work well. More samples tend to yield better approximation at the cost of computation. The overall WHDR on the IIW test split using Nystr{\"o}m approximated pairwise comparison is $17.2$, which is slightly worse than the direct network output ($15.7$).

\begin{figure*}[t!]
\centering
\includegraphics[width=0.95\linewidth]{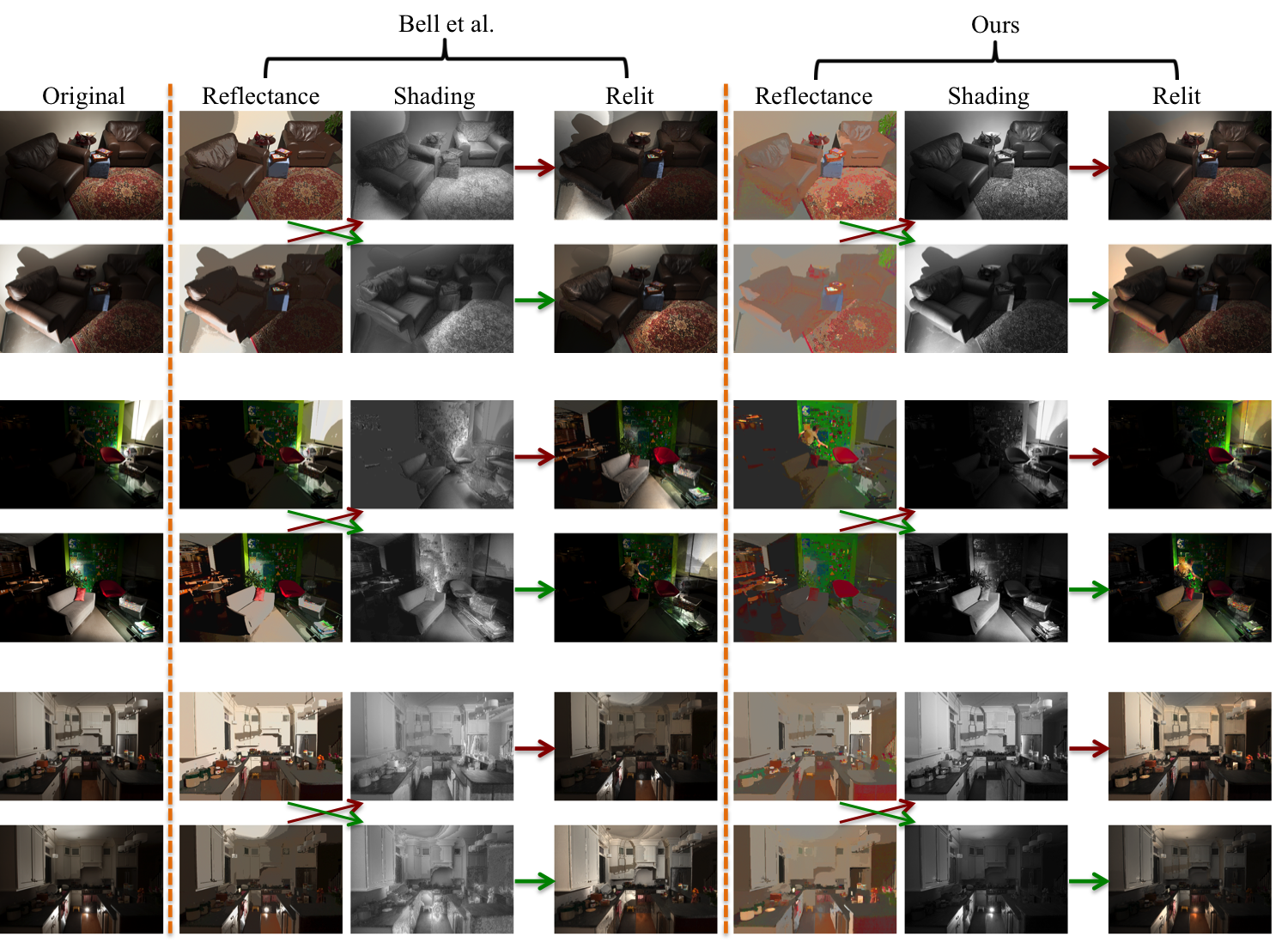}
\vspace{-3mm}
\caption{Comparison of relighting results between Bell~\emph{et al.}~\cite{bbs-iiw-14} and ours on the variable lighting dataset of~\cite{webcam}.
In each row, we construct a relit image from the shading in the same row and the reflectance of the adjacent row. We expect a minimal change in appearance between the original and relit images, since they depict the same scene and thus should share the same reflectance.
Our relighting results tend to reconstruct the target images more closely, which also implies better decomposition performance.
See Sec.~\ref{sec:relight} for more details.}
\label{fig:relight}
\vspace{-3mm}
\end{figure*}

\subsection{Intrinsic image decomposition}
To understand the effect of our reflectance prior on intrinsic image decomposition, we perform an ablation study on several variants of the decomposition framework:

\begin{table}[b]
\centering       
\scalebox{0.9}{
\begin{tabular}{ccc}  
\toprule
 Data source & Original & Augmented \tabularnewline
 Metric & WHDR & Error Rate \tabularnewline
 \midrule
Bell \etal\cite{bbs-iiw-14} & 20.6 & 27.9 \tabularnewline
Chromaticity only & 33.6 &  38.5 \tabularnewline
Chrom. + Our prior & 22.5 & 29.6 \tabularnewline
Chrom. + Our prior + Shading & $\mathbf{19.9}$ & $\mathbf{27.3}$ \tabularnewline
\bottomrule
\end{tabular}
}
\caption{Ablation study on different variants of the decomposition framework. All results are on the test set of IIW.}
\label{tab:err}
\vspace{-3mm}
\end{table}

\paragraph{Chromaticity only} each pixel being assigned to the reflectance label that is most similar in chromaticity. This simple variant achieves WHDR $= 33.6$ on the original annotations, and error rate $= 38.5$ on the augmented data.

\paragraph{Chromaticity + our prior} dense CRF with chromaticity similarity as the unary potential and our reflectance prior as the pairwise potential. This variant greatly improves the performance over using chromaticity only with WHDR $= 22.5$ and error rate $=29.6$ on the original and augmented annotations, respectively, indicating the effectiveness of our reflectance prior.

\paragraph{Chromaticity + our prior + shading} previous variant with additional shading costs from Bell \etal~\cite{bbs-iiw-14}. This variant achieves the best decomposition performance with WHDR $= 19.9$ and error rate $=27.3$. It improves on the decomposition of Bell~\etal both quantitatively and qualitatively.

We visualize our final decomposition output (chrom. + our prior + shading), and compare with Bell~\etal~\cite{bbs-iiw-14} in Figure~\ref{fig:decomp} for examples from both IIW dataset and ILC datasets. In general, our decomposition tends to distinguish between reflectance and shading boundaries better than the baseline, especially under challenging lighting conditions (e.g. examples from the ILC dataset). For instance, for the kitchen scene in the first row of Fig.~\ref{fig:decomp}, Bell~\etal failed to separate the shading layer from the reflectance layer correctly, leading to large shadow boundaries (see cupboards and the floor) left over in the reflectance layer. Similarly for the example in row 6 of Fig.~\ref{fig:decomp}, Bell~\etal failed to recognize that the drastic intensity change on the ceiling and floor is due to illumination from the lamp, whereas our decomposition was able to correctly identify the shadows, and attribute them to the shading layer. However, the hand-crafted reflectance smoothness prior still works more favorably in some cases (e.g. the last row of Fig.~\ref{fig:decomp}).

\begin{figure*}[b!]
\centering
\definecolor{ta3skyblue}{rgb}{0.12549, 0.29020, 0.52941}	%
\definecolor{ta3scarletred}{rgb}{0.64314, 0, 0}			%
\begin{tikzpicture}[inner sep = 0pt, ta3scarletred!50!black, scale=1.445, frame/.style={rectangle, draw, ultra thick},inset/.style={rectangle, draw, ultra thick}, con/.style={densely dashed, thick}]
\node (tsne) at (0,0){\includegraphics[width=.5\linewidth]{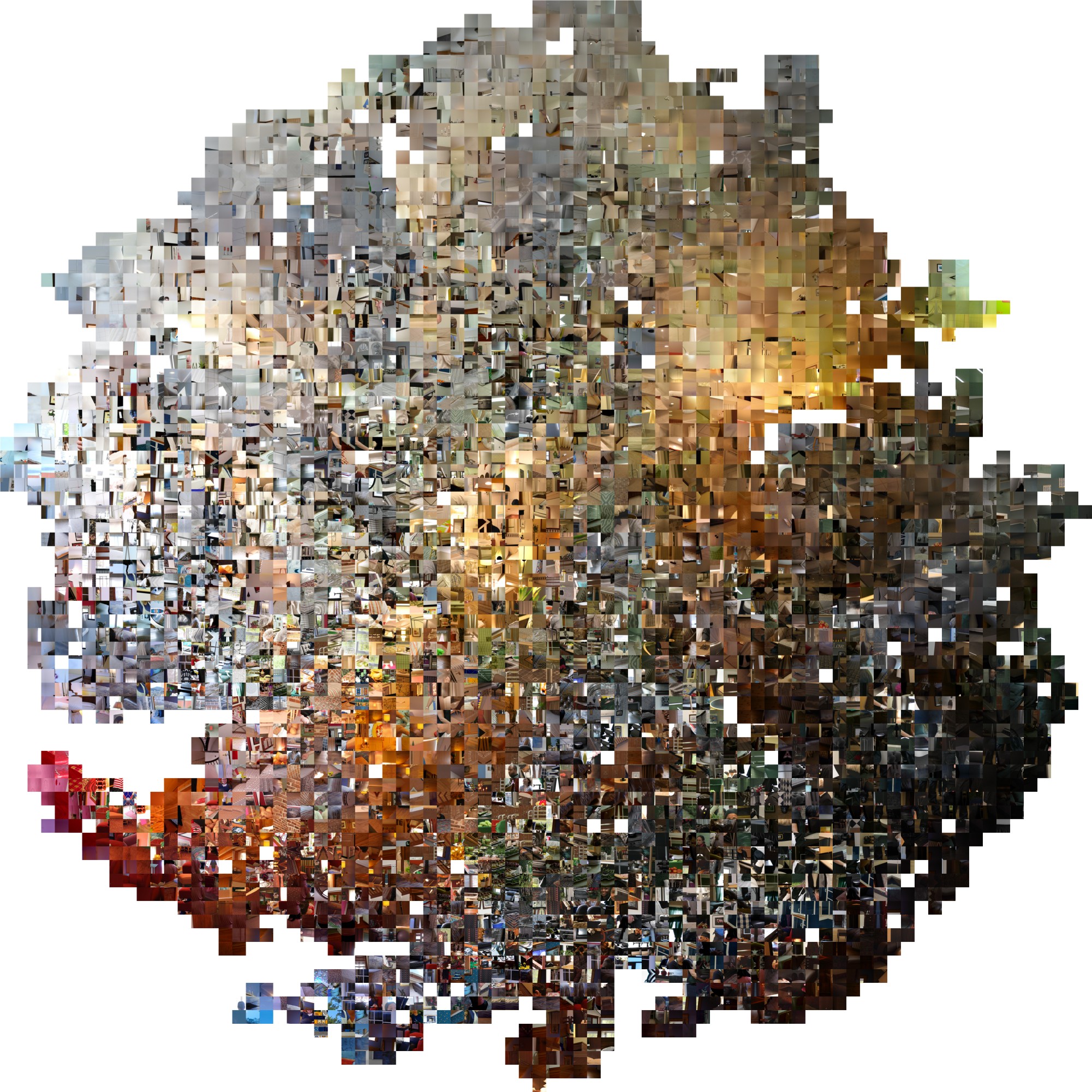}};
\node[inset] (c1) at (-4.5,-2){\includegraphics[width=.25\linewidth]{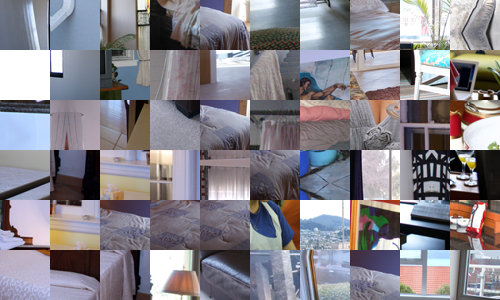}};
\node[frame, fit={(-2.6520, -0.4546) (-1.8943, -0.9093)}] (b1) {};
\node[inset] (c2) at (-4.5,1.4){\includegraphics[width=.15\linewidth]{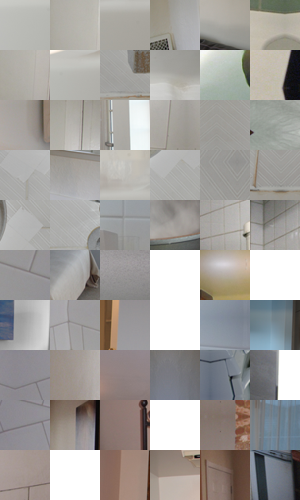}};
\node[frame, fit={(-1.7427, 2.3489) (-1.2881, 1.5912)}] (b2) {};
\node[inset] (c3) at (4.5,-2){\includegraphics[width=.25\linewidth]{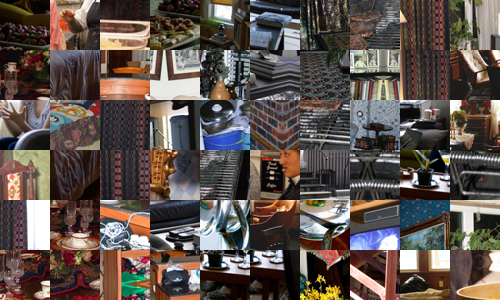}};
\node[frame, fit={(-0.6062, -2.0458) (0.1515, -2.5005)}] (b3) {};
\node[inset] (c4) at (4.5, 2){\includegraphics[width=.25\linewidth]{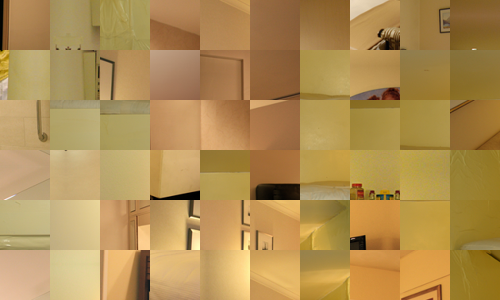}};
\node[frame, fit={(0.8335, 1.4397) (1.5912, 0.9850)}] (b4) {};

\draw[con]   (c1.north west) -- (b1.north west) (c1.south east) -- (b1.south east);
\draw[con]   (c2.north east) -- (b2.north east) (c2.south east) -- (b2.south east);
\draw[con]   (c3.north west) -- (b3.north west) (c3.south west) -- (b3.south west);
\draw[con]   (c4.north west) -- (b4.north west) (c4.south west) -- (b4.south west);
\end{tikzpicture}
\caption{Feature embedding visualized by t-SNE~\cite{tsne}. The learned features are usually highly predictive of surface color (bottom right). More interestingly, our network is also able to coherently group patches based on properties beyond reflectance. For example, the network groups bathroom tiles (top left), wall paper (top right), or cloth surfaces (bottom left), based on material properties or local appearance.}
\label{fig:embed}
\vspace{-3mm}
\end{figure*}
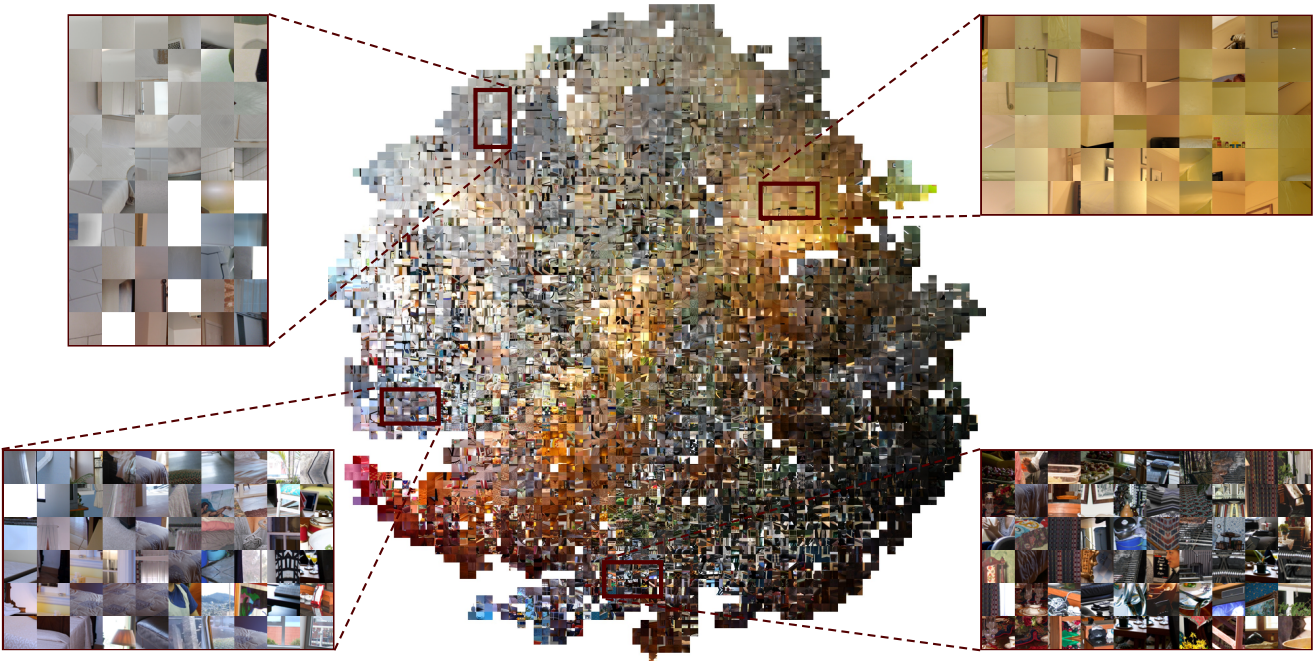

\subsection{Robustness to illumination variation}
\label{sec:relight}
\begin{table}[t] 
\centering       
\scalebox{0.9}{
\begin{tabular}{ccccc}  
\toprule
 & Kitchen & Sofas & Cafe & Mean \tabularnewline
 \midrule
Bell \etal\cite{bbs-iiw-14} & 8.66 & 8.39 & 8.55 & 8.53 \tabularnewline
Ours & $\mathbf{6.93}$ & $\mathbf{6.87}$ & $\mathbf{6.63}$ & $\mathbf{6.81}$ \tabularnewline
\bottomrule
\end{tabular}                                   
}
\vspace{-2mm}
\caption{Mean pixel reconstruction error (MPRE) on three illumination varying sequences ($\times 10^{-4}$). Lower is better.}
\label{tab:mpre}             
\vspace{-4mm}
\end{table}
An ideal reflectance model should be invariant to illumination changes. To measure the degree of illumination invariance, we use image sequences of indoor scenes taken by a stationary camera under different lighting conditions provided by~\cite{webcam}, and perform relighting experiments on decomposition outputs of our method and Bell~\etal. Specifically, given two images $I_A$ and $I_B$ taken from the same scene and their decomposition $I_A = R_A S_A$ and $I_B = R_B S_B$ respectively, perfect decomposition would imply equal reflectance $R_A = R_B$, and the difference between $I_A$ and $I_B$ is entirely explained by the shading/lighting components $S_A$ and $S_B$. In other words, for ideal decompositions, we should be able to relight $R_A$ using $S_B$ to reconstruct $I_B$ (and similarly use $R_B$ and $S_A$ to reconstruct $I_A$). Thus, we propose to use mean pixel reconstruction error (MPRE), $\frac{1}{N^2P}\sum_A\sum_B\|R_AS_B - I_A\|_2$, for measuring illumination invariance, where $N$ is the number of images, and $P$ is the number of pixels per image. %
We report the MPRE results for the three indoor scene sequences in Table~\ref{tab:mpre}, and a qualitative comparison in Fig.~\ref{fig:relight}. We significantly outperform Bell \emph{et al.} both quantitatively and perceptually.

\subsection{Feature visualization}
Finally, we visualize the features learned by our relative reflectance network using the t-SNE algorithm~\cite{tsne}. Specifically, we randomly extract $50,000$ patches from the test set of IIW and find a 2-dimensional embedding of their 64-dimensional Conv4 features. Fig.~\ref{fig:embed} shows this embedding. The overall layout appears to be highly predictive of reflectance (light to dark from top-left to bottom-right). Moreover, it seems to discover some surface or material properties beyond reflectance (see Fig.~\ref{fig:embed} for more details).

\section*{Discussion}
One limitation of our paper is that although the learned reflectance prior accounts for most of the decomposition performance, hand-crafted unaries on chromaticity and shading are still used for achieving state-of-the-art results. However, while it is beyond the scope of this paper, we believe hand-crafted unaries can be replaced by learned unaries (c.f. concurrent work of Narihira~\etal~\cite{takuya-iccv15}).
\noindent\textbf{Acknowledgements}  This research is supported by ONR MURI N000141010934, Intel research grant, and Nvidia hardware donation.

{\small
\bibliographystyle{ieee}
\bibliography{egbib}
}

\end{document}